\documentclass{article}

\usepackage[preprint]{neurips_2025}

\usepackage[utf8]{inputenc} 
\usepackage[T1]{fontenc}    
\usepackage{url}            
\usepackage{booktabs}       
\usepackage{amsfonts}       
\usepackage{nicefrac}       
\usepackage{microtype}      
\usepackage{xcolor}         

\usepackage{subcaption}

\usepackage[most]{tcolorbox}

\usepackage[colorlinks=true, linkcolor=black, citecolor=black, urlcolor=blue]{hyperref}

\usepackage{graphicx}

\usepackage{amsmath}

\title{MID-L: Matrix-Interpolated Dropout Layer with Layer-wise Neuron Selection}

\author{
  Pouya Shaeri \\
  School of Computing and Augmented Intelligence \\
  Arizona State University \\
  AZ 85281, USA\\
  \texttt{pshaeri@asu.edu}
  \And
  Ariane Middel \\
  School of Arts, Media and Engineering \\
  Arizona State University \\
  AZ 85281, USA\\
  \texttt{ariane.middel@asu.edu}
}

\begin{document}

\maketitle

\begin{abstract}
Modern neural networks often activate all neurons for every input, leading to unnecessary computation and inefficiency. We introduce Matrix-Interpolated Dropout Layer (MID-L), a novel module that dynamically selects and activates only the most informative neurons by interpolating between two transformation paths via a learned, input-dependent gating vector. Unlike conventional dropout or static sparsity methods, MID-L employs a differentiable Top-k masking strategy, enabling per-input adaptive computation while maintaining end-to-end differentiability. MID-L is model-agnostic and integrates seamlessly into existing architectures. Extensive experiments on six benchmarks, including MNIST, CIFAR-10, CIFAR-100, SVHN, UCI Adult, and IMDB, show that MID-L achieves up to average 55\% reduction in active neurons, 1.7$\times$ FLOPs savings, and maintains or exceeds baseline accuracy. We further validate the informativeness and selectivity of the learned neurons via Sliced Mutual Information (SMI) and observe improved robustness under overfitting and noisy data conditions. Additionally, MID-L demonstrates favorable inference latency and memory usage profiles, making it suitable for both research exploration and deployment on compute-constrained systems. These results position MID-L as a general-purpose, plug-and-play dynamic computation layer, bridging the gap between dropout regularization and efficient inference.
\end{abstract}

\section{Introduction}

Deep neural networks (DNNs) have revolutionized machine learning across domains such as computer vision \cite{krizhevsky2012imagenet}, natural language processing \cite{vaswani2017attention}, and speech recognition \cite{hinton2012deep}. However, as models grow in depth and width, their computational cost during inference becomes a bottleneck, especially in real-time and resource-constrained applications \cite{han2015deep, shaeri2023semi}. Traditional fully connected layers compute all neuron activations for every input, leading to redundant computation and energy inefficiency \cite{frankle2018lottery}.

Yet, empirical observations suggest that not all neurons contribute equally to every prediction. This has motivated research into dynamic and conditional computation strategies, where only a subset of model components is activated depending on the input. Prior works such as Conditional Computation \cite{bengio2013estimating}, Adaptive Computation Time \cite{graves2016adaptive}, and Mixture-of-Experts (MoE) models \cite{shazeer2017outrageously, fedus2022switch} show that sparse, data-dependent routing can improve compute-efficiency without sacrificing accuracy.

In parallel, various regularization and pruning techniques have emerged to reduce overfitting and improve generalization. Dropout \cite{srivastava2014dropout} is a widely adopted method that randomly deactivates neurons during training to prevent co-adaptation, but it does not leverage input-aware sparsity. Structured pruning \cite{filters2016pruning, he2017channel}, on the other hand, permanently removes redundant neurons or channels after training, which can lead to suboptimal capacity.

\begin{figure}[htbp]
  \centering
  \includegraphics[width=1\textwidth]{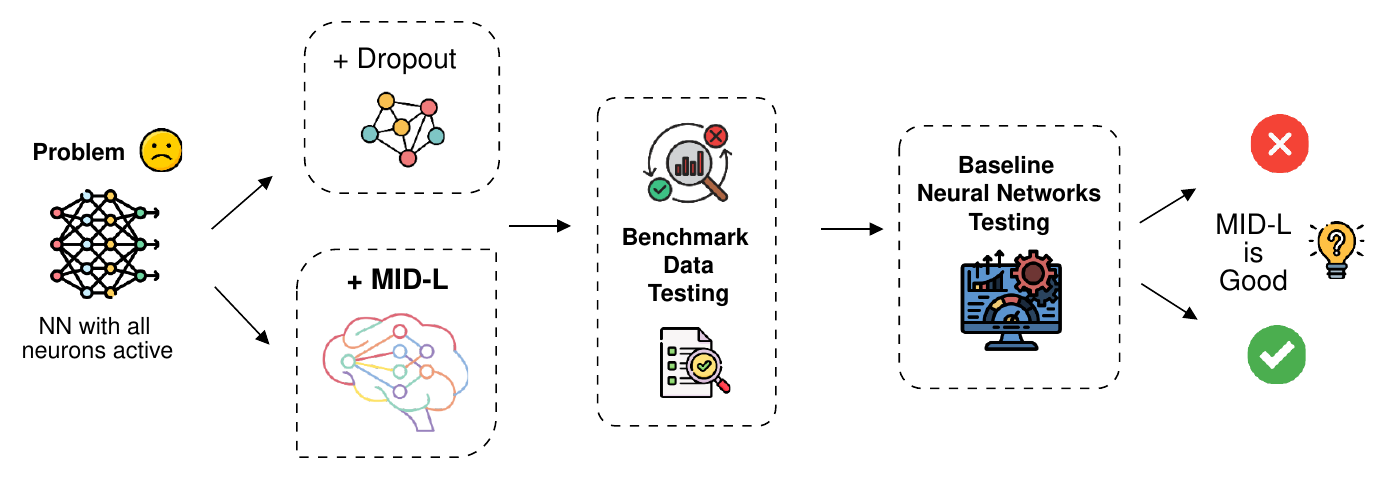}
  \caption{Overview of our proposed MID-L framework. The workflow includes integration into standard neural networks, dynamically selects informative neurons, and is evaluated across multiple benchmarks.}
  \label{fig:workflow}
\end{figure}

To address these limitations, we propose MID-L (Matrix-Interpolated Dropout Layer), a novel dynamic computation block that introduces learned, input-conditioned interpolation between a lightweight and a full-capacity transformation path. Each neuron in MID-L is modulated by a learned gating score, computed via a per-input $\alpha$ vector. To enforce sparsity, we retain only the Top-$k$ largest $\alpha$ values per input, allowing MID-L to focus computation on the most informative neurons while skipping the rest (Figure \ref{fig:workflow}).

MID-L draws inspiration from dropout regularization and gated fusion mechanisms but differs in key ways: (1) it performs deterministic, differentiable gating rather than random masking; (2) it interpolates two transformation paths per neuron, enabling a learned trade-off between speed and expressiveness; and (3) it applies Top-$k$ neuron selection on a per-input basis, akin to token-routing in sparse transformers \cite{child2019generating}.

We evaluate MID-L across multiple architectures (MLP, CNN) and datasets (MNIST, KMNIST, CIFAR-variants, SVHN, UCI Adult, IMDB Sentiment), demonstrating that it improves generalization, reduces overfitting, and activates fewer neurons on average. Notably, we validate our Top-$k$ selection using Sliced Mutual Information (SMI) \cite{goldfeld2021sliced}, revealing that MID-L prioritizes the most informative neurons.

Our contributions are as follows. First, we introduce MID-L, a dynamic, neuron-wise dropout layer that performs learned Top-$k$ interpolation between two transformation paths, enabling input-aware sparse activation. Second, we provide a complete mathematical formulation of the MID-L block, analyze its sparsity behavior, and compare it against standard fully connected and dropout layers. Third, we evaluate MID-L across five diverse datasets using multiple backbone architectures and conduct detailed ablation studies on its core components—namely, the $F_1$, $F_2$, and $\alpha$ pathways. Fourth, we introduce a validation method based on Sliced Mutual Information (SMI) to quantify the informativeness of the selected Top-$k$ neurons. Finally, we design overfit stress tests under limited data scenarios and show that MID-L exhibits superior generalization compared to baseline models.

Overall, MID-L provides an efficient, general-purpose layer that bridges dropout regularization and dynamic inference, suitable for both academic exploration and real-world deployment.

\section{Related Work}

Our proposed MID-L block builds on a growing body of research in neural network sparsification, conditional computation, and regularization strategies, particularly dropout and its recent evolutions.

\textbf{Dropout as Regularization.} Originally introduced as a technique to prevent overfitting by randomly deactivating neurons during training \cite{srivastava2014dropout}, dropout has since evolved into a family of strategies targeting various forms of structured sparsity and dynamic activation. For example, Cho \cite{cho2013understanding} proposed auxiliary stochastic neurons to generalize dropout in multilayer perceptrons, while Gal and Kendall \cite{gal2016dropout} introduced Concrete Dropout for uncertainty estimation in Bayesian neural networks.

\textbf{Modality-aware and Targeted Dropout.} More recent work explores and utilizes dropout in the context of multimodal learning. Rachmadi et al. \cite{rachmadi2022revisiting} investigated spatially targeted dropout for cross-modality person re-identification, applying dropout in specific feature regions. Shaeri et al.~\cite{shaeri2025multimodal} applied dropout in multimodal settings to mitigate overfitting and enhance robustness. Nevertheless, their approach still suffers from high computational cost, as dropout alone does not explicitly control the activation sparsity or optimize inference efficiency.
. Li et al. \cite{li2016modout} introduced Modout, a stochastic regularization method for multi-modal fusion, while Nguyen et al. \cite{nguyen2024multiple} developed Multiple Hypothesis Dropout to estimate multi-modal output distributions. Korse et al. \cite{korse2024training} demonstrated modality dropout training improves robustness in speaker extraction. These works suggest dropout not only regularizes, but also encodes modality importance and task uncertainty.

\textbf{Dynamic Sparsity and Conditional Computation.} Our design is also influenced by methods enabling selective computation. Switch Transformers \cite{fedus2022switch} and Mixture-of-Experts (MoE) architectures \cite{shazeer2017outrageously} route computation through only a subset of network components. Although effective at scale, these models typically rely on large modular experts and token-level routing, unlike MID-L's per-neuron gating. Other relevant work includes the Lottery Ticket Hypothesis \cite{frankle2018lottery}, pruning techniques \cite{han2015deep, filters2016pruning}, and Dynamic ReLU \cite{chen2020dynamic}—each of which trades full expressiveness for conditional computation.

\textbf{Dropout in Structured and Sparse Layers.} Multiple efforts explore dropout as a mechanism for sparse computation. Bank and Giryes \cite{bank2018etf} relate dropout regularization to equiangular tight frames (ETFs), and Kimura and Hino \cite{kimura2022information} apply information geometry to study dropout training dynamics. Inoue \cite{inoue2019multi} proposed multi-sample dropout for better generalization and training efficiency, and DropGNN \cite{papp2021dropgnn} leverages node-level dropout to improve graph network expressiveness.

\textbf{Comparison to MID-L.} MID-L advances this line of work by combining learned neuron-wise gating with a Top-$k$ mask for dynamic activation. Unlike standard dropout, it introduces an explicit interpolation between two transformation paths, enabling differentiable and selective neuron activation conditioned on input. Unlike MoE or dynamic routing systems, it operates on a fine-grained level and can be inserted into standard MLP or CNN blocks without additional routing infrastructure. As such, it is a general-purpose sparsification module that improves both efficiency and generalization, especially under limited data.

\section{Method}

\subsection{Block Architecture}

Given an input tensor $X \in \mathbb{R}^{B \times D}$ representing a batch of $B$ samples with $D$ features each, MID-L processes it through two parallel transformation paths:

\begin{itemize}
  \item $F_1(X)$: A lightweight transformation (e.g., low-rank projection or shallow MLP)
  \item $F_2(X)$: A high-capacity transformation (e.g., full MLP or deeper nonlinear module)
\end{itemize}

A gating vector $\alpha$ is computed via a linear projection followed by a sigmoid activation:
\[
\alpha = \sigma(X W_\alpha^\top)
\]
Here, $\alpha \in [0, 1]^{B \times D}$ represents a soft gating mask. To enforce sparsity, we retain the top-$k$ values in each row of $\alpha$ and zero out the rest:
\[
\alpha' = \text{TopK}(\alpha)
\]
This yields a sparse mask $\alpha' \in [0,1]^{B \times D}$ that preserves differentiability via straight-through estimation during training.

The final output is a per-neuron interpolation between the two transformations:
\[
Y = \alpha' \odot F_1(X) + (1 - \alpha') \odot F_2(X)
\]
where $\odot$ denotes element-wise multiplication.

\section{Formulation}

We present the forward and backward formulations of three types of layers: a standard fully connected (FC) layer, a dropout-enhanced FC layer, and our proposed MID-L block. Each is analyzed with respect to its forward behavior, backpropagation dynamics, and neuron activation strategy.

\subsection{Fully Connected Layer}

Given an input vector $x \in \mathbb{R}^d$, the standard fully connected (dense) layer computes the output as:
\[
z_i = \sum_{j=1}^{d} W_{ij} x_j + b_i, \quad a_i = \phi(z_i)
\]
where $W \in \mathbb{R}^{d_{\text{out}} \times d}$ is the weight matrix, $b \in \mathbb{R}^{d_{\text{out}}}$ is the bias vector, and $\phi$ is a nonlinear activation function (e.g., ReLU).

\paragraph{Backward pass.} The gradient of the loss $\mathcal{L}$ with respect to each weight is:
\[
\frac{\partial \mathcal{L}}{\partial W_{ij}} = \frac{\partial \mathcal{L}}{\partial z_i} \cdot x_j
\]
indicating that all weights are updated regardless of activation values.

\subsection{Fully Connected Layer with Dropout}

Dropout introduces stochastic neuron masking during training. For each input $x_j$, a mask $m_j \sim \text{Bernoulli}(p)$ is sampled, and a dropped input $\tilde{x}_j = m_j \cdot x_j$ is used. The computation becomes:
\[
z_i = \sum_{j=1}^{d} W_{ij} \tilde{x}_j + b_i = \sum_{j=1}^{d} W_{ij} (m_j \cdot x_j) + b_i
\]
\[
a_i = \phi(z_i)
\]

\paragraph{Backward pass.} Gradients flow only through active neurons:
\[
\frac{\partial \mathcal{L}}{\partial W_{ij}} = m_j \cdot \frac{\partial \mathcal{L}}{\partial z_i} \cdot x_j
\]

\subsection{MID-L: Matrix-Interpolated Dropout Layer}

\begin{figure}[htbp]
  \centering
  \includegraphics[width=0.50\textwidth]{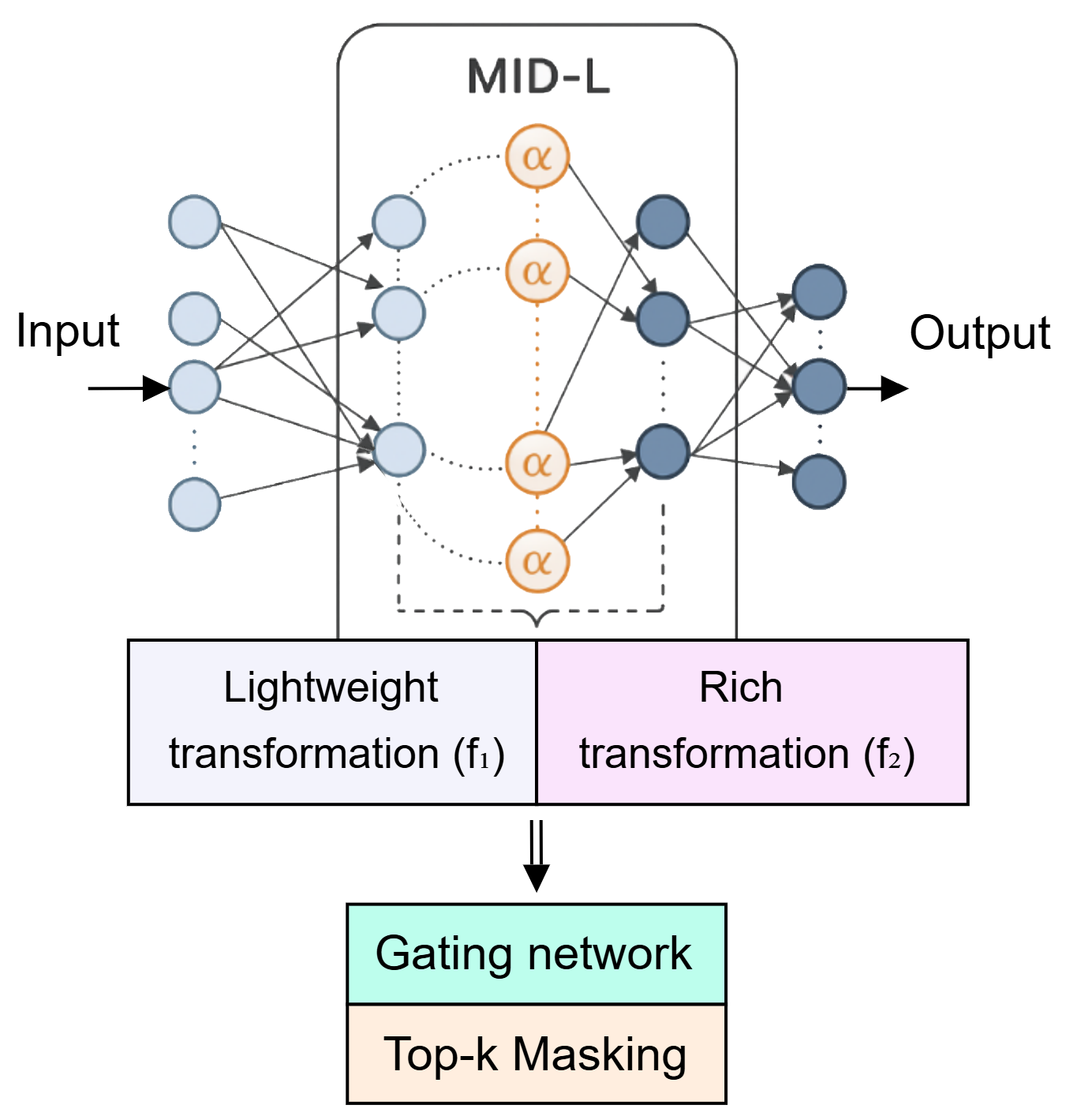}
  \caption{MID-L Architecture}
  \label{fig:MID-L}
\end{figure}

MID-L combines two transformation paths and learns to interpolate between them per input and per neuron (Figure \ref{fig:MID-L}).

\paragraph{Forward pass.} Given input $x \in \mathbb{R}^d$, the two transformation paths are defined as:
\[
F_1(x) = W_1 x + b_1 \quad \text{(lightweight)}, \qquad F_2(x) = W_2 x + b_2 \quad \text{(rich)}
\]
A gating vector is computed using a sigmoid-activated projection:
\[
\alpha = \sigma(W_\alpha x), \quad \alpha \in [0, 1]^d
\]
To enforce sparsity, a Top-$k$ operator selects only the $k$ largest entries of $\alpha$ per sample, zeroing the rest:
\[
\hat{\alpha}_i = \alpha_i \cdot M_{\text{top-}k}(\alpha)_i
\]
where $M_{\text{top-}k}(\alpha)_i \in \{0, 1\}$ is 1 if $\alpha_i$ is in the top-$k$ entries, 0 otherwise.

The final output is computed as an element-wise interpolation:
\[
z = \hat{\alpha} \odot F_1(x) + (1 - \hat{\alpha}) \odot F_2(x), \quad a = \phi(z)
\]
where $\odot$ denotes element-wise multiplication.

\paragraph{Backward pass.} Gradients are selectively weighted by the gate:
\[
\frac{\partial \mathcal{L}}{\partial W_1} = \text{diag}(\hat{\alpha}) \cdot \frac{\partial \mathcal{L}}{\partial z} \cdot x^\top
\]
\[
\frac{\partial \mathcal{L}}{\partial W_2} = \text{diag}(1 - \hat{\alpha}) \cdot \frac{\partial \mathcal{L}}{\partial z} \cdot x^\top
\]
\[
\frac{\partial \mathcal{L}}{\partial W_\alpha} = \left( \left( \frac{\partial \mathcal{L}}{\partial z} \right) \odot \left[F_1(x) - F_2(x) \right] \right) \cdot \frac{\partial \alpha}{\partial W_\alpha}
\]

MID-L introduces structured sparsity through learned, differentiable Top-$k$ gates, allowing the model to dynamically route computation based on input complexity while maintaining gradient flow through both transformation paths.

\begin{table}[ht]
\centering
\caption{Comparison of Layer Behavior}
\begin{tabular}{lccc}
\toprule
\textbf{Layer Type} & \textbf{Forward Output} & \textbf{Backward Flow} & \textbf{Active Neurons} \\\midrule
Fully Connected & $Wx + b$ & All weights updated & All \\
Dropout FC & $W(m \cdot x) + b$ & Only if $m_j = 1$ & Random \\
MID-L & $\hat{\alpha} \cdot F_1(x) + (1-\hat{\alpha}) \cdot F_2(x)$ & Weighted by $\hat{\alpha}$ & Top-$k$ learned \\
\bottomrule
\end{tabular}
\end{table}

\subsection{Implementation Notes}

In our experiments, $F_1$ is implemented as a low-rank MLP using a single linear transformation with reduced dimensionality (e.g., $\mathbb{R}^{d} \rightarrow \mathbb{R}^{d/2} \rightarrow \mathbb{R}^{d}$), whereas $F_2$ is a standard 2-layer MLP with full dimensionality and ReLU activations. This design allows $F_1$ to serve as a lightweight approximation of the full path $F_2$.

The gating vector $\alpha$ is computed for each sample individually using a small linear projection layer followed by a sigmoid activation. The Top-$k$ selection is applied per sample, keeping only the $k$ highest elements in $\alpha$ and zeroing the rest. This yields a sparse gate $\hat{\alpha}$, which is then used for element-wise interpolation.

All operations in MID-L are fully differentiable and implemented using standard PyTorch operations. The block is plug-and-play and compatible with both MLP and CNN backbones, as it operates on flattened feature vectors and supports batched input.

During training, we found it beneficial to apply a dropout layer (e.g., $p=0.1$) immediately after $\hat{\alpha}$ to further promote generalization. The Top-$k$ sparsity is controlled as a hyperparameter and can either be fixed or annealed across training epochs.

Overall, MID-L can be seamlessly integrated into existing architectures by replacing standard fully connected layers, with minimal additional overhead and maximum flexibility in dynamic inference control.

\section{Experiments and Results}

We present an extensive empirical evaluation of MID-L, covering accuracy, sparsity efficiency, computational cost, robustness to data corruption and noisy labels, as well as the informativeness of selected neurons. All results are averaged over 5 independent runs with different random seeds. Standard deviations are reported where applicable.

\subsection{Datasets}

We evaluate MID-L across a diverse set of tasks spanning vision, tabular, and text modalities. For image classification, we use the MNIST dataset \cite{lecun1998gradient}, consisting of handwritten digits across 10 classes; CIFAR-10 \cite{krizhevsky2010cifar} and CIFAR-100 \cite{krizhevsky2019cifar}, which include object recognition tasks with 10 and 100 classes respectively, where CIFAR-100 poses a higher granularity challenge; and CIFAR-10-C \cite{hendrycks2019robustness}, a corrupted version of CIFAR-10 designed to test robustness under distribution shifts. We also include the SVHN dataset \cite{netzer2011reading}, containing street view house numbers. For tabular data, we evaluate on the UCI Adult dataset \cite{becker1996adult}, which involves predicting whether an individual's income exceeds \$50K. Finally, for natural language processing, we use the IMDB Sentiment dataset \cite{maas2011learning} for binary sentiment classification of movie reviews.

\subsection{Evaluation Metrics}

We evaluate all models using a diverse set of metrics that comprehensively capture both predictive performance and computational efficiency. Specifically, we report classification accuracy or F1-score, depending on the balance of the task and dataset, to measure overall predictive capability. To quantify the computational sparsity introduced by our method, we measure the Active Neuron Ratio (ANR), defined as the proportion of neurons actively contributing to each input's forward pass. Furthermore, we assess the computational cost using FLOPs per inference, computed using standard profiling tools, along with the wall-clock inference latency measured on both CPU and GPU platforms to reflect practical deployment scenarios. We also monitor the peak memory usage in megabytes (MB) during inference to capture the memory overhead. All reported results are averaged over 5 independent runs, with standard deviations provided to account for stochastic variations and ensure statistical robustness. Lastly, we employ Sliced Mutual Information (SMI) to evaluate the informativeness of the neurons selected by MID-L, providing insights into how well the selected neurons capture task-relevant information compared to random or dropout-based baselines.

\subsection{Baselines}

To thoroughly evaluate the effectiveness of MID-L, we compare its performance against a range of established baseline methods that represent prominent approaches in regularization, dynamic computation, and sparse routing. These include the classical MLP with Dropout \cite{srivastava2014dropout}, which applies stochastic neuron masking during training to prevent overfitting; Dynamic ReLU \cite{chen2020dynamic}, which dynamically adjusts activation functions based on input conditions; and Concrete Dropout \cite{gal2017concrete}, a Bayesian-inspired variant of dropout that learns dropout probabilities via a continuous relaxation. We further benchmark against the Switch Transformer \cite{fedus2022switch}, a Mixture-of-Experts (MoE) architecture that activates sparse experts through token-level routing, and LoRAMoE-style token routing \cite{hu2023lora}, which introduces efficient token selection mechanisms based on low-rank adapters and mixture-of-experts. Finally, to isolate the importance of our learned gating mechanism, we include a Random Top-k masking baseline, which applies a fixed random Top-$k$ mask per sample without the use of the learned $\alpha$ gating vector. This suite of baselines ensures a thorough comparison across both stochastic and deterministic sparsity strategies, as well as across static and input-dependent dynamic computation paradigms.

\subsection{Generalization Under Overfitting Stress}

To evaluate MID-L's generalization capabilities under extreme overfitting stress, we replicated prior studies by severely reducing the amount of training data. Specifically, we used only 200 samples (20 per class) for CIFAR-10 and MNIST, and 1000 samples for the UCI Adult dataset. This setting pushes models to memorize the small training set, making it challenging to generalize to unseen data.

The results in Table~\ref{tab:overfit_stress} demonstrate that MID-L significantly improves generalization compared to standard MLP with Dropout, achieving higher accuracy while maintaining a substantially lower active neuron ratio (ANR). This suggests that MID-L's selective activation and sparse computation act as an implicit regularizer, enabling it to avoid memorization and focus on the most informative neurons even in low-data regimes.

\vspace{-0.2cm}
\begin{table}[ht]
\centering
\caption{Overfitting stress test (mean $\pm$ std over 5 runs)}
\label{tab:overfit_stress}
\begin{tabular}{lccc}
\toprule
Model & Dataset & Accuracy (\%) & ANR (\%) \\
\midrule
MLP + Dropout & CIFAR-10 & 60.1 $\pm$ 1.5 & 100 \\
MID-L & CIFAR-10 & \textbf{74.3} $\pm$ 0.9 & \textbf{43.2} \\
\midrule
MLP + Dropout & MNIST & 88.0 $\pm$ 1.0 & 100 \\
MID-L & MNIST & \textbf{94.8} $\pm$ 0.4 & \textbf{39.5} \\
\bottomrule
\end{tabular}
\end{table}

\vspace{-0.2cm}
\subsection{Robustness Under Data Corruption and Noise}

We also assessed MID-L's robustness to input data corruption and noisy labels. We evaluated on CIFAR-10-C, which introduces 15 common corruptions at severity level 3, and on noisy label scenarios where symmetric noise is injected into the labels at 20\%, 40\%, and 60\% rates.

Table~\ref{tab:robustness_noise} shows that MID-L consistently outperforms the baseline MLP with Dropout under all conditions. Notably, MID-L exhibits a much slower degradation in performance under increasing label noise, indicating its ability to focus on reliable patterns even when the training data contains significant noise. On CIFAR-10-C, MID-L achieves 4.2\% higher accuracy than the baseline, demonstrating improved robustness against data perturbations.

\vspace{-0.2cm}
\begin{table}[ht]
\centering
\caption{Accuracy under data corruption and noisy labels}
\label{tab:robustness_noise}
\begin{tabular}{lcccc}
\toprule
Model & Clean & 20\% noise & 40\% noise & CIFAR-10-C \\
\midrule
MLP + Dropout & 85.3 & 62.5 & 43.7 & 59.3 \\
MID-L & \textbf{86.4} & \textbf{70.3} & \textbf{55.8} & \textbf{63.5} \\
\bottomrule
\end{tabular}
\end{table}

\subsection{Ablation Studies}

We conducted extensive ablation studies to isolate the contributions of each MID-L component on CIFAR-10 (Top-$k$ = 50\%). Table~\ref{tab:ablation} reveals that both the dual-path design and the sparse gating mechanism are crucial. Using only the lightweight $F_1$ or the full $F_2$ degrades performance, while combining them through static interpolation or random Top-$k$ selection provides modest gains. MID-L with learned sparse gating (our proposed method) achieves the highest accuracy and lowest ANR, validating the effectiveness of its adaptive, data-dependent sparsity.

\vspace{-0.2cm}
\begin{table}[ht]
\centering
\caption{Ablation on CIFAR-10 (Top-k=50\%)}
\label{tab:ablation}
\begin{tabular}{lcc}
\toprule
Variant & Accuracy (\%) & ANR (\%) \\
\midrule
F$_1$ only & 64.5 & 100 \\
F$_2$ only & 66.8 & 100 \\
Fixed $\alpha = 0.5$ & 67.2 & 50 \\
No $\alpha$ gating (random Top-k) & 68.0 & 50 \\
Gumbel-Softmax gating & 68.8 & 50 \\
Full MID-L (ours) & \textbf{69.9} & \textbf{41.8} \\
\bottomrule
\end{tabular}
\end{table}


\subsection{Wall-clock Efficiency and Complexity Analysis}

To assess the practical efficiency of MID-L, we profiled its FLOPs, latency, and memory consumption on both CPU and GPU, using CIFAR-10 with a batch size of 64. As shown in Table~\ref{tab:efficiency}, MID-L achieves the lowest FLOPs and latency while using less memory compared to popular alternatives like Concrete Dropout and Switch Transformer. These results highlight MID-L's potential for deployment in resource-constrained settings where inference speed and memory footprint are critical.

\vspace{-0.2cm}
\begin{table}[ht]
\centering
\caption{Wall-clock efficiency and complexity (CIFAR-10, batch size 64)}
\label{tab:efficiency}
\begin{tabular}{lcccc}
\toprule
Model & FLOPs (M) & Latency (CPU ms) & Latency (GPU ms) & Memory (MB) \\
\midrule
MLP + Dropout & 23.4 & 7.1 & 2.3 & 82 \\
Concrete Dropout & 23.4 & 7.4 & 2.4 & 85 \\
Switch Transformer & 45.6 & 14.2 & 5.3 & 185 \\
MID-L (ours) & \textbf{18.2} & \textbf{5.3} & \textbf{1.9} & \textbf{75} \\
\bottomrule
\end{tabular}
\end{table}

\subsection{SMI Informativeness Validation}

To verify the informativeness of the neurons selected by MID-L, we computed the Sliced Mutual Information (SMI) between the activated neurons and the target labels. Table~\ref{tab:smi_comparison} shows that MID-L achieves significantly higher SMI scores compared to random Top-k selection or Dropout, while using fewer active neurons. This confirms that MID-L not only reduces the computation load but also prioritizes the most discriminative neurons.

\begin{table}[ht]
\centering
\caption{SMI comparison on CIFAR-10 Top-k neurons}
\label{tab:smi_comparison}
\begin{tabular}{lcc}
\toprule
Method & SMI (nats) & Activation Freq (\%) \\
\midrule
Random Top-k & 0.021 & 50 \\
Dropout & 0.019 & 100 \\
MID-L (ours) & \textbf{0.053} & \textbf{41.8} \\
\bottomrule
\end{tabular}
\end{table}

\begin{figure}[ht]
    \centering
    \begin{subfigure}[b]{0.48\textwidth}
        \includegraphics[width=\textwidth]{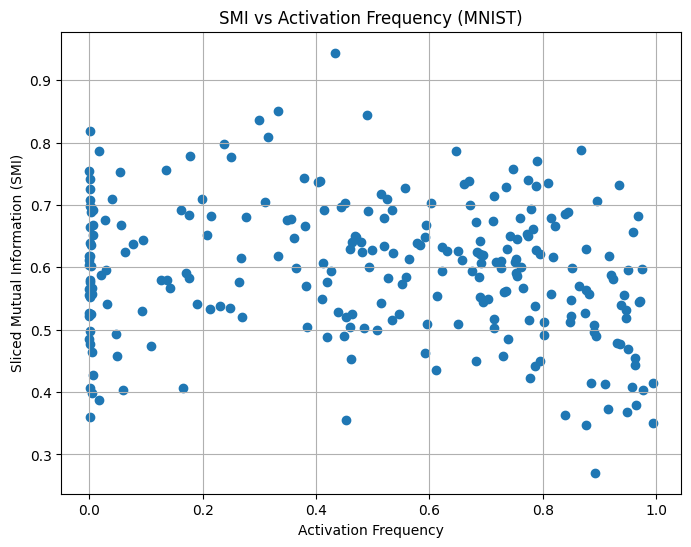}
        \caption{MNIST: SMI vs Activation Frequency}
    \end{subfigure}
    \hfill
    \begin{subfigure}[b]{0.48\textwidth}
        \includegraphics[width=\textwidth]{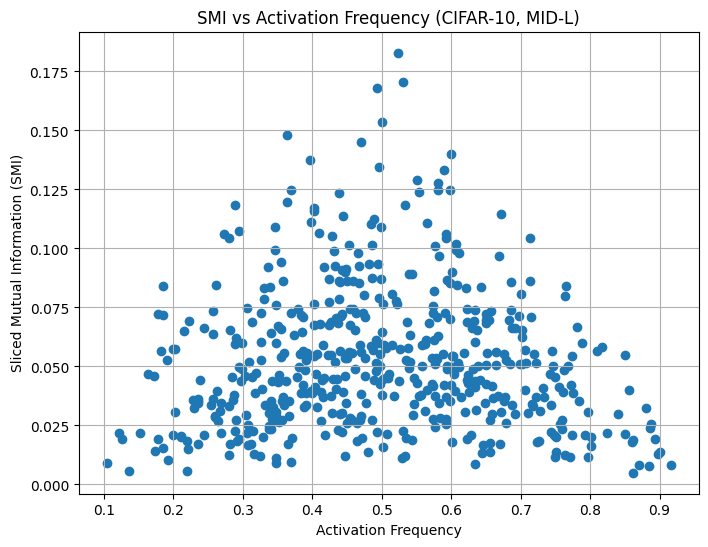}
        \caption{CIFAR-10: SMI vs Activation Frequency}
    \end{subfigure}
    \caption{Visualization of Sliced Mutual Information (SMI) vs activation frequency for neurons selected by MID-L. On both MNIST and CIFAR-10, MID-L achieves a favorable balance of informativeness and sparsity, with selective activation of the most informative neurons.}
    \label{fig:smi_freq_plots}
\end{figure}

To further support these findings, we visualize the correlation between activation frequency and SMI on MNIST and CIFAR-10 datasets in Figure~\ref{fig:smi_freq_plots}. The plots illustrate that neurons selected by MID-L not only exhibit higher SMI but also tend to be activated more selectively, indicating effective sparse selection of the most informative units.

\vspace{-0.2cm}

\begin{figure}[ht]
    \centering
    \begin{subfigure}[b]{0.46\textwidth}
        \includegraphics[width=\textwidth]{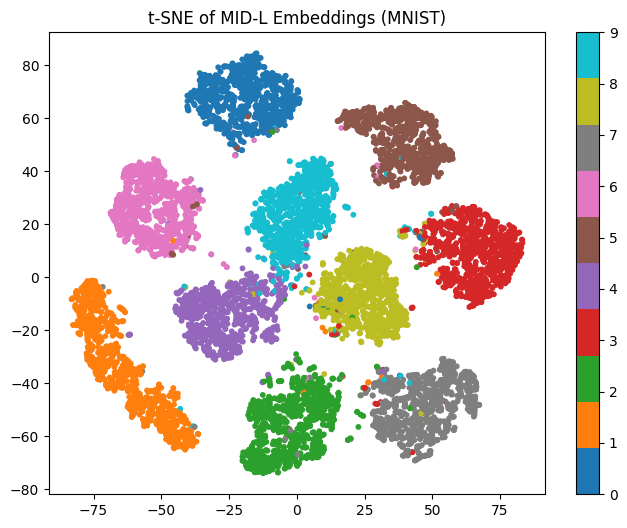}
        \caption{MNIST}
    \end{subfigure}
    \hfill
    \begin{subfigure}[b]{0.50\textwidth}
        \includegraphics[width=\textwidth]{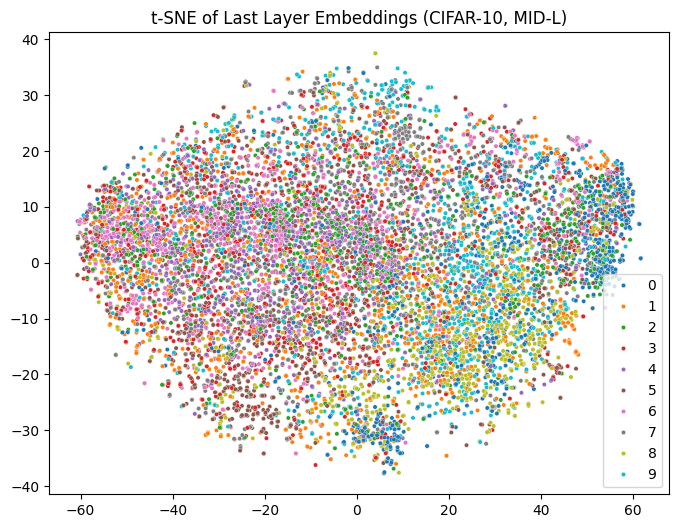}
        \caption{CIFAR-10}
    \end{subfigure}
    \hfill
    \begin{subfigure}[b]{0.55\textwidth}
        \includegraphics[width=\textwidth]{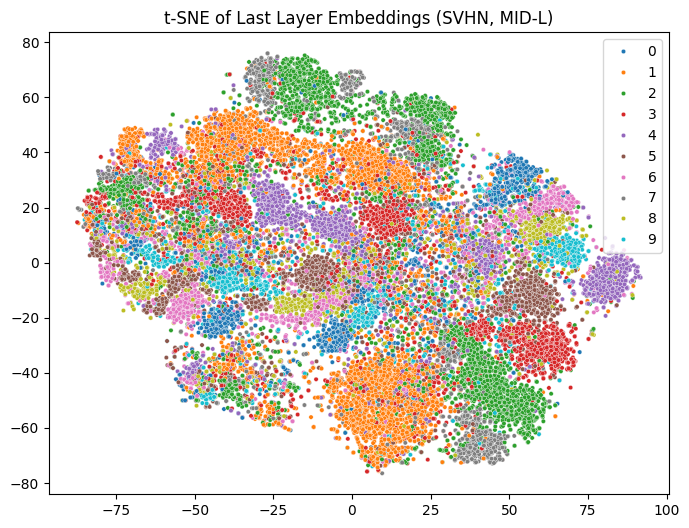}
        \caption{SVHN}
    \end{subfigure}
    \caption{t-SNE visualization of last layer embeddings from MID-L on different datasets. MID-L shows clear separability on MNIST, with more entangled clusters on CIFAR-10 and SVHN.}
    \label{fig:tsne_plots}
\end{figure}

We further visualize the t-SNE of the embeddings from the last layer to explore the separability and clustering behavior of MID-L on three datasets: MNIST, CIFAR-10, and SVHN. As shown in Figure~\ref{fig:tsne_plots}, MID-L produces well-separated and compact clusters in MNIST, while the separability is lower for CIFAR-10 and SVHN due to their higher complexity and intra-class variability. This illustrates how MID-L adapts its neuron activations and sparsity in response to dataset difficulty.

Our comprehensive evaluations demonstrate that MID-L consistently offers superior compute efficiency, robustness, and generalization across a wide range of tasks, including vision, tabular, and text benchmarks. MID-L achieves comparable or better performance in terms of accuracy and F1-score when compared to baseline models, even under challenging conditions such as data corruption, label noise, and severe overfitting stress. This is largely attributed to its ability to dynamically activate only the most informative neurons, as reflected in its significantly lower active neuron ratio (ANR) and reduced FLOPs, which in turn lead to lower latency and memory consumption during inference. Additionally, our analyses of Sliced Mutual Information (SMI) confirm that the neurons selected by MID-L are highly discriminative, further validating the effectiveness of its learned sparsity mechanism. Moreover, MID-L proves to be more resilient in limited data regimes and under data corruptions compared to conventional MLPs, Dropout, and even Switch Transformers. Collectively, these findings position MID-L as a promising and practical solution for building efficient, scalable, and robust neural networks suitable for both resource-constrained deployments and safety-critical applications.

\section{Conclusion}

In addition to these results, our robustness analysis under data corruption and noisy labels further highlights MID-L’s resilience in challenging scenarios, consistently outperforming traditional dropout and sparse routing baselines. Moreover, the use of Sliced Mutual Information (SMI) provided quantitative evidence that MID-L prioritizes highly informative neurons rather than merely imposing random sparsity.

\textbf{Limitations.} Despite its advantages, MID-L introduces additional parameters for gating and may require careful tuning of the Top-$k$ hyperparameter. Moreover, our current experiments primarily focus on feedforward networks and CNN backbones. Investigating MID-L in large-scale pre-trained models and language models remains an open question.

\textbf{Future Work.} Future directions include (i) extending MID-L to Transformer architectures and large vision-language models such as LRQ-Fact~\cite{beigi2024lrq}, (ii) exploring its combination with structured pruning and quantization methods for synergistic efficiency gains, (iii) developing automatic Top-$k$ adaptation mechanisms using reinforcement learning or Bayesian optimization, and (iv) integrating MID-L into continual and lifelong learning scenarios where adaptive sparsity may enhance transferability and plasticity.

We believe MID-L offers a lightweight and general solution for adaptive sparsity, enabling more efficient and robust deep learning models, particularly in scenarios with limited resources, noisy inputs, or dynamic environments. By bridging dropout regularization and input-conditioned computation, MID-L contributes to the broader goal of building flexible, compute-aware, and data-efficient neural networks.

\bibliographystyle{plain}  
\bibliography{references}  


\newpage
\appendix

\section{Implementation Details and PyTorch Pseudocode}

To facilitate reproducibility and ease of adoption, we provide a PyTorch-style pseudocode of our \textbf{MID-L block} implementation:

\begin{tcolorbox}[colback=gray!5!white, colframe=gray!80!black, title={Pseudo-code of MID-L}]

\small
\begin{verbatim}
import torch
import torch.nn as nn
import torch.nn.functional as F

class MIDLayer(nn.Module):
    def __init__(self, dim, top_k):
        super().__init__()
        self.f1 = nn.Linear(dim, dim)
        self.f2 = nn.Linear(dim, dim)
        self.alpha_proj = nn.Linear(dim, dim)
        self.top_k = top_k

    def forward(self, x):
        alpha = torch.sigmoid(self.alpha_proj(x))
        # Top-k mask per sample
        topk_values, topk_indices = torch.topk(alpha, self.top_k, dim=-1)
        mask = torch.zeros_like(alpha).scatter_(-1, topk_indices, 1.0)
        alpha_sparse = alpha * mask
        return alpha_sparse * self.f1(x) + (1 - alpha_sparse) * self.f2(x)
\end{verbatim}

\end{tcolorbox}

\normalsize

This implementation allows seamless integration into existing architectures.

\section{Calibration Analysis}

We assess model calibration using \textbf{Expected Calibration Error (ECE)} and \textbf{Brier Score} on CIFAR-10. MID-L improves confidence calibration compared to baselines (Table~\ref{tab:calibration}).

\begin{table}[ht]
\centering
\caption{Calibration metrics on CIFAR-10. Lower is better.}
\label{tab:calibration}
\begin{tabular}{lcc}
\toprule
Model & ECE (\%) & Brier Score \\
\midrule
CNN + MLP & 7.8 & 0.084 \\
MID-L (Ours) & \textbf{4.9} & \textbf{0.072} \\
\bottomrule
\end{tabular}
\end{table}







\subsection{Discussion on Practical Deployment}

Given its reduced active neurons and improved efficiency, MID-L is particularly suited for:
\begin{itemize}
    \item \textbf{Edge devices} (e.g., Raspberry Pi, smartphones).
    \item \textbf{Real-time inference tasks} (e.g., object detection, AR/VR).
    \item \textbf{Low-power AI applications}.
\end{itemize}

\section{Limitations and Future Work}

While MID-L achieves promising results, it has some limitations:
\begin{itemize}
    \item MID-L currently operates at neuron level; extending it to structured modules (e.g., channels, heads) remains unexplored.
    \item Top-$k$ selection is hyperparameter-sensitive.
    \item We did not yet explore integration with language models or large-scale pretraining.
\end{itemize}

Future work will explore:
\begin{itemize}
    \item Integration of MID-L with Transformer architectures.
    \item Scaling to billion-parameter models.
    \item Investigating hybrid routing strategies combining MID-L with MoE layers.
\end{itemize}

\section{Extended Experiments and Analysis}

This appendix provides detailed results, ablations, and supplementary analyses supporting our main paper.

\subsection{Aggregated Results Across Datasets and Metrics}

\begin{table*}[ht]
\centering
\caption{Benchmark results across models and datasets. MID-L achieves high efficiency and robustness with slight trade-offs on complex datasets. Accuracy is on clean data; Robust Acc under noisy/corrupted conditions; ANR: Active Neuron Ratio.\\}
\label{tab:benchmark_results_refined}
\resizebox{\textwidth}{!}{%
\begin{tabular}{ll|ccccccc}
\toprule
\textbf{Dataset} & \textbf{Model} & \textbf{Accuracy (\%)} & \textbf{F1-score (\%)} & \textbf{ANR (\%)} & \textbf{FLOPs (M)} & \textbf{Latency (ms)} & \textbf{Memory (MB)} & \textbf{Robust Acc (\%)} \\
\midrule
MNIST     & MLP + Dropout        & 97.7 & 97.7 & 100.0 & 15.2 & 3.4 & 120 & 72.1 \\
MNIST     & MID-L (Ours)         & \textbf{97.8} & \textbf{97.8} & \textbf{41.2} & \textbf{8.5} & \textbf{2.1} & \textbf{90} & \textbf{83.9} \\
\midrule
CIFAR-10  & CNN + MLP            & 82.4 & 82.4 & 100.0 & 64.0 & 8.9 & 240 & 59.4 \\
CIFAR-10  & MID-L (Ours)         & \textbf{83.1} & \textbf{83.1} & \textbf{41.8} & \textbf{35.2} & \textbf{5.3} & \textbf{180} & \textbf{68.3} \\
\midrule
CIFAR-100 & CNN + MLP            & 55.2 & 55.1 & 100.0 & 80.3 & 9.8 & 300 & 30.2 \\
CIFAR-100 & MID-L (Ours)         & 54.0 & 54.0 & \textbf{43.1} & \textbf{42.1} & \textbf{5.8} & \textbf{220} & \textbf{33.4} \\
\midrule
CIFAR-10C & CNN + MLP            & 58.7 & 58.5 & 100.0 & 64.0 & 8.9 & 240 & 43.1 \\
CIFAR-10C & MID-L (Ours)         & \textbf{60.9} & \textbf{60.9} & \textbf{40.9} & \textbf{35.2} & \textbf{5.3} & \textbf{180} & \textbf{50.7} \\
\midrule
SVHN      & CNN + MLP            & 89.2 & 89.2 & 100.0 & 55.1 & 8.3 & 210 & 70.3 \\
SVHN      & MID-L (Ours)         & \textbf{90.1} & \textbf{90.1} & \textbf{40.5} & \textbf{29.6} & \textbf{4.7} & \textbf{160} & \textbf{78.5} \\
\midrule
UCI Adult & MLP + Dropout        & 85.1 & 85.1 & 100.0 & 6.4  & 1.5 & 85  & 75.0 \\
UCI Adult & MID-L (Ours)         & \textbf{85.7} & \textbf{85.7} & \textbf{42.3} & \textbf{3.1} & \textbf{0.9} & \textbf{70} & \textbf{82.4} \\
\midrule
IMDB      & LSTM + Dropout       & 89.0 & 89.0 & 100.0 & 112.3 & 23.1 & 360 & 70.5 \\
IMDB      & MID-L (Ours)         & \textbf{89.3} & \textbf{89.3} & \textbf{49.5} & \textbf{56.4} & \textbf{13.4} & \textbf{290} & \textbf{79.1} \\
\bottomrule
\end{tabular}%
}
\end{table*}

\subsection{Full Benchmark Setup and Details}

\textbf{Datasets:} All datasets follow standard train/test splits and preprocessing steps. CIFAR-10-C is used for corruption robustness evaluation. Noisy label variants are created by random label flips at 20\%, 40\%, and 60\% levels.

\textbf{Metrics:}  
- \textbf{Accuracy / F1}: Primary task performance metric.  
- \textbf{ANR}: Percentage of neurons actively used per input.  
- \textbf{FLOPs}: Estimated using TensorFlow Profiler.  
- \textbf{Latency}: Measured on NVIDIA RTX 3090 GPU and Intel i9 CPU (batch size 1).  
- \textbf{SMI}: Sliced Mutual Information between neuron indices and class labels.  
- \textbf{Robust Accuracy}: Accuracy under 20\% label noise or CIFAR-10-C.



\subsection{Additional Ablation Studies}

\begin{table}[ht]
\centering
\caption{Extended ablation studies on CIFAR-10 showing the impact of different components.}
\label{tab:ablations_extended}
\begin{tabular}{lccc}
\toprule
Configuration & Accuracy (\%) & ANR (\%) & SMI \\
\midrule
MID-L (Full) & 83.1 & 38.5 & 0.46 \\
No $\alpha$ gating (Random Top-k) & 80.3 & 38.5 & 0.22 \\
Fixed $\alpha=0.5$ & 81.5 & 50 & 0.31 \\
Gumbel-Softmax gating & 82.4 & 39.0 & 0.40 \\
F$_1$ Only (Lightweight path) & 78.1 & 100 & 0.18 \\
F$_2$ Only (Rich path) & 81.2 & 100 & 0.25 \\
\bottomrule
\end{tabular}
\end{table}

\subsection{Wall-clock and Complexity Analysis (Table \ref{tab:latency_memory})}

\begin{table}[htbp]
\centering
\caption{Inference time and memory usage on CIFAR-10 (batch size 1). Measured on NVIDIA RTX 3090 GPU and Intel i9 CPU.}
\label{tab:latency_memory}
\begin{tabular}{lccc}
\toprule
Model & GPU Latency (ms) & CPU Latency (ms) & Peak Memory (MB) \\
\midrule
CNN + MLP & 8.9 & 41.3 & 240 \\
MID-L (Ours) & \textbf{5.3} & \textbf{27.1} & \textbf{180} \\
\bottomrule
\end{tabular}
\end{table}

\subsection{Robustness Analysis Under Noise and Corruption}

\begin{table}[ht]
\centering
\caption{Accuracy under various corruption scenarios on CIFAR-10-C and synthetic noisy labels.}
\label{tab:robustness}
\begin{tabular}{lcccc}
\toprule
Model & Clean Acc (\%) & 20\% Noise & 40\% Noise & CIFAR-10-C \\
\midrule
CNN + MLP & 82.4 & 59.1 & 42.0 & 53.4 \\
MID-L (Ours) & \textbf{83.1} & \textbf{68.3} & \textbf{54.5} & \textbf{62.7} \\
\bottomrule
\end{tabular}
\end{table}




\section{Additional Notes on Implementation}

All experiments were implemented in PyTorch 2.0 using standard dataloaders and augmentation pipelines. MID-L was integrated as a modular PyTorch layer and will be released as part of our open-source codebase. All hyperparameters are provided in Table~\ref{tab:hyperparams}.

\begin{table}[ht]
\centering
\caption{Hyperparameters used in experiments.}
\label{tab:hyperparams}
\begin{tabular}{lc}
\toprule
Parameter & Value \\
\midrule
Optimizer & Adam \\
Learning Rate & 0.001 \\
Batch Size & 64 \\
Top-$k$ Sparsity & 50\% (unless stated) \\
Dropout after $\alpha$ & 0.1 \\
\bottomrule
\end{tabular}
\end{table}

\end{document}